\pdfoutput=1
\documentclass[runningheads]{llncs}
\usepackage{graphicx}
\usepackage{amsmath,amssymb} 
\usepackage{color}
\usepackage{booktabs}
\usepackage{marvosym}
\usepackage{fancyhdr}

\begin{document}

\newcommand{\point}{
    \raise0.7ex\hbox{.}
    }


\pagestyle{headings}

\mainmatter

\title{Signature of Geometric Centroids for 3D Local Shape Description and Partial Shape Matching} 
\titlerunning{SGC for 3D Local Shape Description and Partial Shape Matching} 
\authorrunning{Keke Tang, Peng Song, and Xiaoping Chen} 

\author{Keke Tang,  Peng Song$^{\text{(\Letter)}}$  and Xiaoping Chen} 
\institute{University of Science and Technology of China 
} 

\maketitle

\thispagestyle{fancy}
\fancyhead{}
\chead{\small{In Proceedings of the Asian Conference on Computer Vision (ACCV'2016)}}
\renewcommand{\headrulewidth}{0pt}       


\begin{abstract}
Depth scans acquired from different views may contain nuisances such as noise, occlusion, and varying point density.
We propose a novel {\em Signature of Geometric Centroids} descriptor, supporting direct shape matching on the scans, without requiring any preprocessing such as scan denoising or converting into a mesh.
First, we construct the descriptor by voxelizing the local shape within a uniquely defined local reference frame and concatenating geometric centroid and point density features extracted from each voxel.
Second, we compare two descriptors by employing only corresponding voxels that are both non-empty, thus supporting matching incomplete local shape such as those close to scan boundary.
Third, we propose a descriptor saliency measure and compute it from a descriptor-graph to improve shape matching performance.
We demonstrate the descriptor's robustness and effectiveness for shape matching by comparing it with three state-of-the-art descriptors, and applying it to object/scene reconstruction and 3D object recognition.
\end{abstract}

\section{Introduction}
\label{sec:intro}


The recent development in depth sensing devices offers a convenient and flexible way to acquire depth scans of an object or a scene that represent their partial shapes.
In practice, we need to register these scans into a common coordinate system to better understand the object's or scene's geometry~\cite{Aiger-2008-4PointsCongruent} or compare known object models with these scans for 3D object recognition~\cite{Bariya-2010-ScaleHierarchical}.
All these applications require solving the partial shape matching problem~\cite{Donoser-2009-OuterContours,Rodola-2016-Partial}.


Depth scans (i.e., 3D point clouds) lack topology information of the shape and usually contain noise, holes, and/or varying point density.
To facilitate partial shape matching, one common way is to convert the point cloud into a mesh to remove the noise and fill the holes, and then perform shape matching on the mesh instead~\cite{Mian-2006-Registration,Wu-2010-IsometryInvariant,Guo-2013-RotProjection,Song-2015-LocalVoxelizer}.
Although this conversion simplifies the matching process, it brings several drawbacks.
First, original partial shape could be modified and/or downsampled by the conversion, e.g., when smoothing the depth scan for denoising.
Second, the mesh topology generated by the conversion could be different from the real one such as incorrectly filled holes, misleading the shape matching.


Therefore, other researchers seek to perform shape matching directly on the point cloud data.
This is generally achieved by representing and matching the scans using local shape descriptors.
Although existing descriptors~\cite{Johnson-1997-SpinImages,Frome-2004-3DShapeContext,Tombari-2010-SHOT,Guo-2015-TriSI} work well on clean depth scans, they have difficulties dealing with original scans acquired under various conditions such as occlusion, clutter, and varying lighting.
This is because these descriptors are sensitive to noise and/or varying point density due to their encoded shape features such as point density~\cite{Johnson-1997-SpinImages,Frome-2004-3DShapeContext} and surface normals~\cite{Tombari-2010-SHOT}, or are sensitive to scan boundary and holes due to their descriptor comparison scheme that is based on the vector distance~\cite{Tombari-2010-SHOT,Guo-2015-TriSI}.


To address above limitations, we propose a {\em Signature of Geometric Centroids} (SGC) descriptor for partial shape matching with three novel components:
\begin{itemize}
\item
{\em A Robust Descriptor.} \
We construct the SGC descriptor by voxelizing the local shape within a uniquely defined local reference frame (LRF) and concatenating the geometric centroid and point density features extracted from each non-empty voxel.
Thanks to the extracted shape features, our descriptor is robust against noise and varying point density.

\item
{\em A Descriptor Comparison Scheme.} \
Rather than simply computing the Euclidean distance between two descriptors, we compute a similarity score between two descriptors based on comparing the extracted features from corresponding voxels that are both non-empty.
By this, the comparison scheme supports shape matching between local shape that are incomplete.

\item
{\em Descriptor Saliency for Shape Matching.} \
Different from keypoint detection~\cite{Guo-2014-3DRecogSurvey} that identifies distinct points locally on a single scan/model, we propose descriptor saliency to measure distinctiveness of SGC descriptors across all input scans and compute it from a descriptor-graph.
Guided by the descriptor saliency, we improve shape matching performance by intentionally selecting distinct descriptors to find corresponding feature points.
\end{itemize}


We evaluate the robustness of SGC against various nuisances including scan noise, varying point density, distance to scan boundary, occlusion, and the effectiveness of using SGC and descriptor saliency for partial shape matching.
Experimental results show that SGC outperforms three start-of-the-art descriptors (i.e., spin image~\cite{Johnson-1997-SpinImages}, 3D shape context~\cite{Frome-2004-3DShapeContext}, and signature of histograms of orientations (SHOT)~\cite{Tombari-2010-SHOT}) on publicly available datasets.
We further apply SGC to two typical applications of partial shape matching, i.e., object/scene reconstruction and 3D object recognition, to demonstrate its usefulness in practice.


\section{Related Work}
\label{sec:related}


\noindent
{\textbf{Shape Matching.}} \
Shape matching aims at finding correspondences between complete or partial models by comparing their geometries.
Many shape matching approaches apply global shape descriptors to characterize the whole shape, for example, using Reeb graphs~\cite{Hilaga-2001-ReebGraph} or skeleton graphs~\cite{Chao-2011-SkeletonGraph} for articulated objects and shape distributions~\cite{Osada-2002-ShapeDistribution} for rigid objects.
However, depth scans acquired from each single view usually have significant missing data.
Matching these partial shapes is a difficult task because, before computing the correspondences of the shapes, we first need to find the common portions among them~\cite{Aiger-2008-4PointsCongruent}.
This requires a careful design of local shape descriptors~\cite{Guo-2016-DescriptorEvaluation} that are less sensitive to occlusion.


\noindent
{\textbf{Local Shape Descriptors.}} \
Local shape descriptors can be classified as low- and high-dimensional, according to the richness of encoded local shape information.
Low-dimensional descriptors such as surface curvature~\cite{Gal-2006-SalientGeometric} and surface hashes~\cite{Albarelli-2010-SurfaceHashes}, are easy to compute, store, and compare, yet have limited descriptive ability.
Compared with them, high-dimensional descriptors provide a fairly detailed description of the local shape around a surface point.
We classify high-dimensional descriptors into three classes according to their attached LRF~\cite{Petrelli-2011-LRFRepeatability}.


\noindent
{\em{Descriptors without an LRF.}} \
Early local shape descriptors are generated by directly accumulating some geometric attributes into a histogram, without building an LRF.
Hetzel et al.~\cite{Hetzel-2001-FeatureHistogram} represented local shape patches by encoding three local shape features (i.e., pixel depth, surface normals, and curvatures) into a multi-dimensional histogram.
Yamany et al.~\cite{Yamany-2002-SurfaceSignatures} described local shape around a feature point by generating a signature image that captures surface curvatures seen from that point.
Kokkinos et al.~\cite{Kokkinos-2012-IntrinsicShapeContext} generated an intrinsic shape context descriptor by shooting geodesic outwards from a keypoint to chart the local surface and creating a 2D histogram of features defined on the chart.

Due to the missing of an LRF, the correspondence built by matching the descriptors is limited to the point spatial position only.
Thus, to match two scans by estimating a rigid transform, at least three pairs of corresponding points need to be found, making the space of searching corresponding points large.


\noindent
{\em{Descriptors with a non-unique LRF.}} \
Researchers later attached an LRF for local shape descriptors to enrich the correspondence with spatial orientation.
By this, two scans can be matched by finding a single pair of corresponding points using the descriptors and estimating the transform based on aligning associated LRFs.
However, since the attached LRF is not unique, a further disambiguation process is required for the generated transform.

Johnson et al.~\cite{Johnson-1999-SpinImages} proposed a spin image descriptor by spinning a 2D image about the normal of a feature point and summing up the number of points that fall into the bins of that image.
Frome et al.~\cite{Frome-2004-3DShapeContext} proposed a 3D shape context (3DSC) descriptor by generating a 3D histogram of accumulated points within a partitioned spherical volume centered at a feature point and aligned with the feature normal.
Mian et al.~\cite{Mian-2006-Registration} proposed a 3D tensor descriptor by constructing an LRF from a pair of oriented points and encoding the intersected surface area into a multidimensional table.
Zhong~\cite{Zhong-2009-IntrinsicSignature} proposed intrinsic shape signatures by improving~\cite{Frome-2004-3DShapeContext} based on a different partitioning of the 3D spherical volume and a new definition of LRF with ambiguity.


\noindent
{\em{Descriptors with a unique LRF.}} \
Recently, researchers constructed a unique LRF from the local shape around a feature point and further describe the local shape relative to the LRF.
Thanks to the unique LRF, the transform to match two scans can be uniquely defined based on aligning corresponding LRFs.

Tombari et al.~\cite{Tombari-2010-SHOT} proposed a SHOT descriptor by concatenating local histograms of surface normals defined on each bin of a partitioned spherical volume aligned with a unique LRF.
Guo et al.~\cite{Guo-2013-RotProjection} constructed a RoPS descriptor by rotationally projecting the neighboring points of a feature point onto 2D planes and calculating a set of statistics within a unique LRF.
Guo et al.~\cite{Guo-2015-TriSI} later generated three signatures representing the point distribution in three cylindrical coordinate systems and concatenated and compressed these signatures into a Tri-Spin-Image descriptor.
Song and Chen~\cite{Song-2015-LocalVoxelizer} developed a local voxelizer descriptor by voxelizing local shape within a unique LRF and concatenating an intersected surface area feature in each voxel, and applied it to surface registration~\cite{Song-2015-Registration}.


SGC is also constructed within a unique LRF.
Compared with above descriptors, the geometric centroid feature that we extract for constructing the descriptor is more robust against noise and varying point density.
Moreover, our descriptor comparison scheme supports matching local shape that is close to the scan boundary.
By this, SGC is more robust for shape matching on point cloud data than state-of-the-art descriptors~\cite{Johnson-1997-SpinImages,Frome-2004-3DShapeContext,Tombari-2010-SHOT}, see Section~\ref{sec:results} for the comparisons.


\section{Signature of Geometric Centroids Descriptor}
\label{sec:descriptor}

This section presents the method to construct an SGC descriptor for the local shape (i.e., support) around a feature point $p$, a scheme to compare a pair of SGC descriptors, and the parameters tuned for generating SGC descriptors.


\subsection{LRF Construction}
Given a feature point $p$ on a scan and a radius $r$, a local support is defined by intersecting the scan with a sphere centered at $p$ with radius $r$.
Taking this support as input, we construct a unique LRF based on principal component analysis (PCA) on the support by using the approach in~\cite{Tombari-2010-SHOT}, see Figure~\ref{fig:desc_constru}(a).
When the normal of $p$ is available, we further improve the disambiguation of LRF axes by enforcing the principal axis
associated with the smallest eigenvalue (i.e., the blue axis in Figure~\ref{fig:desc_constru}(a)) to be consistent with the normal~\cite{Song-2015-LocalVoxelizer}.


\subsection{SGC Construction}
Given the unique LRF, a general way to construct a descriptor is to partition a support into bins, extract shape features from each bin, and concatenate the values representing the shape features into a descriptor vector (or a histogram).


\noindent
{\textbf{Partition the Support.}} \
Given a support $\mathrm{S_p}$ around a feature point $p$, there are three typical approaches to partition $\mathrm{S_p}$ into small local patches.
The first one is to partition the bounding spherical volume of $\mathrm{S_p}$ into girds evenly~\cite{Tombari-2010-SHOT} or logarithmically~\cite{Frome-2004-3DShapeContext} along azimuth, elevation and radial dimensions.
The second one is to partition the angular space of the spherical volume into relatively homogeneously distributed bins~\cite{Zhong-2009-IntrinsicSignature}.
However, the bins generated by these two approaches have varying sizes, which need to be compensated when constructing a descriptor.
In addition, the irregular shape of these bins complicates the segmentation of local shape within each bin for extracting local shape features.

\begin{figure}[!t]
  \centering
  \includegraphics[width=0.88\textwidth]{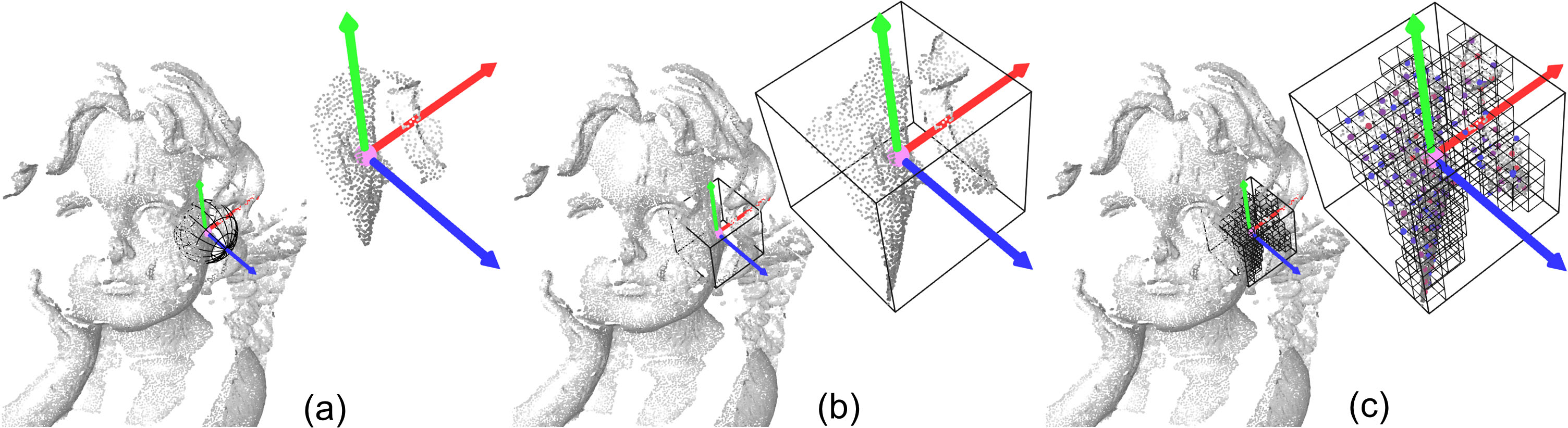}\\
  \caption{Constructing an SGC descriptor.
  (a) Construct a unique LRF from a spherical support centered at a feature point (in pink);
  (b) segment a cubical support centered at the feature point and aligned with the LRF;
  (c) voxelize the support and extract centroid features from non-empty voxels;
  the centroid color indicates point density in the voxel, where small and large densities are colored in blue and red respectively.}
  \label{fig:desc_constru}
\end{figure}

The third approach is to construct a bounding cubical volume of $\mathrm{S_p}$ that is aligned with the LRF and partition the cubical volume into regular bins (i.e., voxels)~\cite{Song-2015-LocalVoxelizer}.
These regular bins simplify the extraction of local shape features and thus the descriptor construction.
Therefore, we employ the third approach to partition $\mathrm{S_p}$ for constructing the SGC descriptor, see Figure~\ref{fig:desc_constru}(b\&c).
Note that the edges of the cubical volume have a length of $2R$, where $R \geq r$.

\noindent
{\textbf{Extract Bin Features.}} \
Due to the missing of topology information, point clouds have limited types of shape features that can be extracted, e.g., surface normal feature in SHOT~\cite{Tombari-2010-SHOT} and point density feature in 3DSC~\cite{Frome-2004-3DShapeContext}.
This paper proposes extracting a geometric centroid feature from each non-empty voxel for constructing SGC due to following reasons.
First, centroid is an integral feature~\cite{Pottmann-2009-Integral}, thus can be more robust against noise and varying point density.
Second, centroid can be computed simply by averaging the positions of all points staying within a voxel.
Note that we do not realize any existing work that employs centroid features for constructing a usable descriptor.


\noindent
{\textbf{Construct the Descriptor.}} \
We divide the cubical volume evenly into $K\times K \times K$ bins (i.e., voxels) with the same size, see Figure~\ref{fig:desc_constru}(c).
For each voxel $\mathrm{V_i}$, we identify all $N_i$ points staying within the voxel and then calculate the centroid ($X_i,Y_i,Z_i$) for the points.
Note that, the position of the centroid is relative to the minimum corner of $\mathrm{V_i}$ in the LRF.
We save the extracted feature as ($X_i,Y_i,Z_i,N_i$) for non-empty voxels, and (0,0,0,0) for empty ones.
An SGC descriptor is generated by concatenating all these values assigned for each voxel.
The dimension of an SGC descriptor saved in this way is $4 \times K\times K \times K$.

Thanks to the unique LRF, the three positional values of $\mathrm{V_i}$'s centroid ($X_i,Y_i,Z_i$) can be compressed into a single value using $C_i = (Z_i \times L+Y_i)\times L+X_i$, where $L=2R$ denotes the edge length of $\mathrm{V_i}$.
By this, we compress the dimension of the descriptor to $2 \times K\times K \times K$, saving 50\% storage space.


\subsection{Comparing SGC Descriptors}

Ideally, SGC descriptors generated for two corresponding points in different scans should be exactly the same.
However, due to variance of sampling, noise and occlusion, the two descriptors usually have a certain amount of difference.
Unlike existing approaches that compare descriptors by computing their Euclidean distance~\cite{Tombari-2010-SHOT,Guo-2013-RotProjection,Song-2015-LocalVoxelizer}, we develop a new scheme for comparing two SGC descriptors.

When constructing an SGC descriptor, most of the voxels are likely to be empty (see again Figure~\ref{fig:desc_constru}(c)).
We classify each pair of corresponding voxels into three cases:
1) empty voxel vs empty voxel; 2) non-empty voxel vs empty voxel; and 3) non-empty voxel vs non-empty voxel.
In all three cases, only case 3 should contribute to computing a similarity score between two descriptors.
Thus, to compare two SGC descriptors quantitatively, we propose to accumulate a similarity score for every pair of corresponding voxels that are both non-empty.

In detail, we denote two SGC descriptors as $\mathrm{D_m}$ and $\mathrm{D_n}$.
The similarity between the $i$-th voxel of $\mathrm{D_m}$, $\mathrm{V_m^i}$, and the $i$-th voxel of $\mathrm{D_n}$, $\mathrm{V_n^i}$, is defined as:
\begin{equation}
s(\mathrm{V_m^i},\mathrm{V_n^i}) \ = \ \left\{ \begin{array}{ll}
\ln{ \frac {N_m^i N_n^i}{{\lVert C_m^i-C_n^i \rVert}^2 + \epsilon } } \mathrm{,} &
\mathrm{for} \ N_m^i > 0 \ and \ N_n^i > 0
\\
0 &
\mathrm{for} \ N_m^i = 0 \ or \ N_n^i = 0
\end{array}
\right.
\label{eq:origin_similarity}
\end{equation}

\noindent
where $N_m^i$ and $N_n^i$ represent the number of points in $\mathrm{V_m^i}$ and $\mathrm{V_n^i}$ respectively, while
$C_m^i$ and $C_n^i$ represent the centroid of $\mathrm{V_m^i}$ and $\mathrm{V_n^i}$ respectively.
Here we directly employ the number of points in each voxel to represent its point density as all voxels have the same size.
The formula can be explained as follows.
Whenever $\mathrm{V_m^i}$ and/or $\mathrm{V_n^i}$ are empty (i.e., $N_m^i=0$ or $N_n^i=0$), $s(\mathrm{V_m^i},\mathrm{V_n^i}) = 0$.
Otherwise, when two corresponding voxels contain similar local shape, their centroids should be close to each other, making $s(\mathrm{V_m^i},\mathrm{V_n^i})$ large.
When $N_m^i$ and/or $N_n^i$ are large, $s(\mathrm{V_m^i},\mathrm{V_n^i})$ is large also as the estimated centroid(s) are more accurate.
By this, the formula encourages to find matches based on denser parts of input scans when the scans are irregularly sampled.

The overall similarity score between $\mathrm{D_m}$ and $\mathrm{D_n}$ can be obtained by accumulating the similarity value for every pair of corresponding voxels:
\begin{equation}
S(\mathrm{D_m}, \mathrm{D_n}) = \sum\limits_{i=1}^{K \times K \times K} s(\mathrm{V_m^i},\mathrm{V_n^i})
\label{eq:desc_compare}
\end{equation}


\subsection{SGC Generation Parameters}
The SGC descriptor has two generation parameters: (i) the support radius $R$; and (ii) the voxel grid resolution $K$.
According to our experiments, we choose $R=20~pr$ as a tradeoff between the descriptiveness and sensitivity to occlusion, where $pr$ denotes the point cloud resolution (i.e., average shortest distance among neighboring points in the scan).
And we choose $K=8$ as a tradeoff between the descriptiveness and efficiency since a larger $K$ increases the descriptiveness and computational cost simultaneously.
Note that in these experiments, we let the LRF and the descriptor have the same support radius, i.e., $r=R$.


\begin{figure*}[!b]
  \centering
  \includegraphics[width=0.95\textwidth]{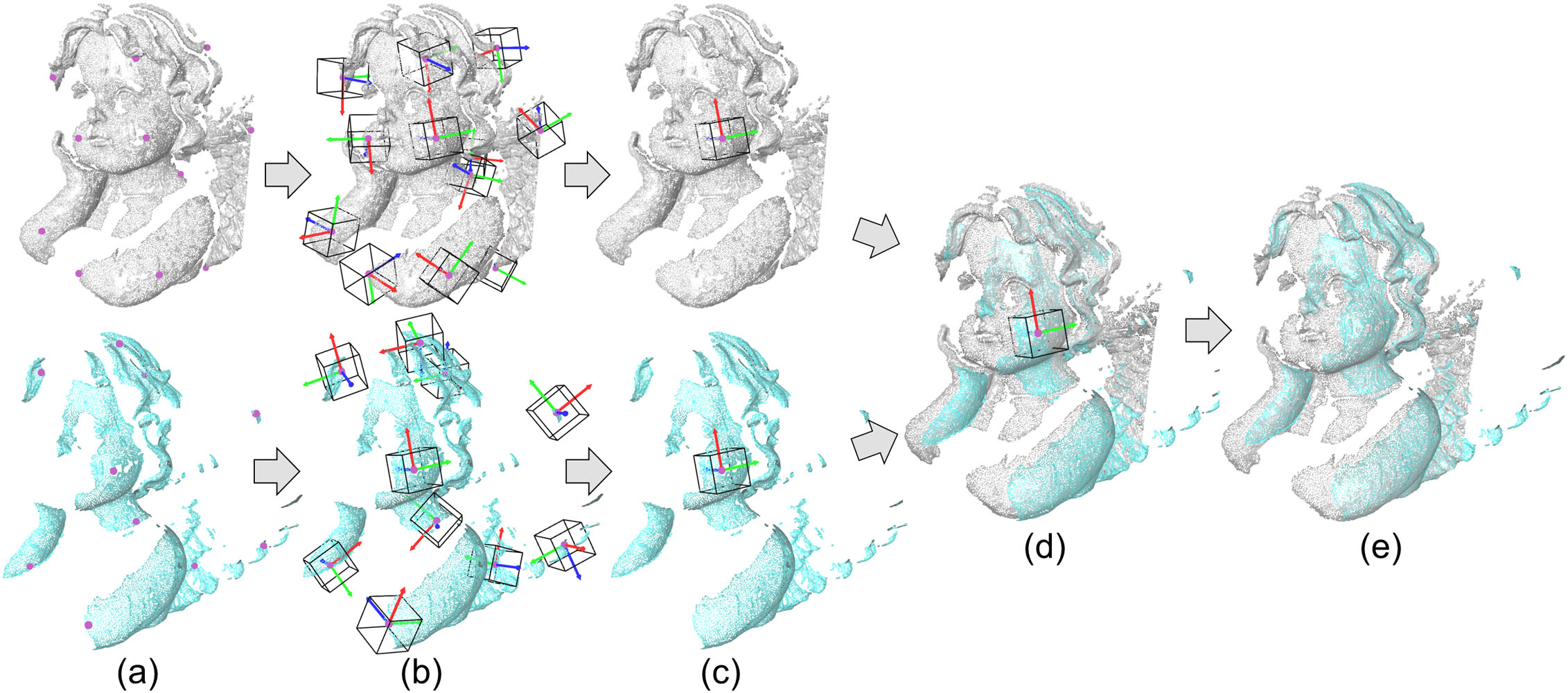}\\
  \caption{Matching two scans using SGC descriptors:
  (a) sampled feature points (in purple) on two input scans (only part of samples are shown for clarity);
  (b) calculated LRFs and descriptors;
  (c) a pair of matched descriptors;
  (d) match the two scans based on aligning the associated LRFs; and
  (e) refine the scan alignment using ICP.}
  \label{fig:match_pipeline}
\end{figure*}

\section{Partial Shape Matching using SGC}
\label{sec:match}

In this section, we describe the general pipeline to match two scans using SGC descriptors and propose a descriptor saliency measure for improving shape matching performance.
We also highlight the advantage of using SGC descriptors for matching supports that are close to scan boundary.


\subsection{General Shape Matching Pipeline}
\label{subsec:match_pipeline}
Given a data scan $\mathrm{S_d}$ and a reference scan $\mathrm{S_r}$, the goal of shape matching between $\mathrm{S_d}$ and $\mathrm{S_r}$ is to find a rigid transform on $\mathrm{S_d}$ to align it with $\mathrm{S_r}$.
By employing the SGC descriptors, we can find such a transform with following steps:

\noindent
{\em {1) Represent Scans with SGC Descriptors.}} \
We first conduct a uniform sampling on each of $\mathrm{S_d}$ and $\mathrm{S_r}$ to generate $M$ feature points that cover the whole scan surface.
Next, for each feature point $p$, we construct the LRF and SGC descriptor for the support around $p$.
By this, we represent each of $\mathrm{S_d}$ and $\mathrm{S_r}$ with $M$ descriptor vectors and the corresponding LRFs, see Figure~\ref{fig:match_pipeline}(a\&b).

\noindent
{\em {2) Generate Transform Candidates.}} \
When a point on $\mathrm{S_d}$ corresponds to another point on $\mathrm{S_r}$, their associated SGC descriptors should be similar to each other.
Hence, we compare each feature descriptor of $\mathrm{S_d}$ with each feature descriptor of $\mathrm{S_r}$ by calculating a similarity score using Eq.~\ref{eq:desc_compare}.
A feature point on $\mathrm{S_d}$ and its closest feature point on $\mathrm{S_r}$ are considered as a match if the similarity score is higher than a threshold.
Each match generates a rigid transform candidate (i.e., a $4\times4$ transformation matrix) by aligning the associated LRFs.

\noindent
{\em {3) Select the Optimal Transform.}} \
By matching the descriptors of $\mathrm{S_d}$ and $\mathrm{S_r}$, we obtain a number of candidate transforms.
We sort these transforms based on the descriptor similarity score and then pick the top five candidates with the highest scores.
We apply each of the five selected transforms on $\mathrm{S_d}$ to align it with $\mathrm{S_r}$.
We evaluate the transform by computing a scan overlap ratio.
We first find all point-to-point correspondences by checking if the distance between a point on transformed $\mathrm{S_d}$ and a point on $\mathrm{S_r}$ is sufficiently small, and further compute the overlap ratio as the number of corresponding points divided by the total number of points in $\mathrm{S_d}$ or $\mathrm{S_r}$ (smaller one).
We select the transform that ensures the largest overlap ratio as the optimal one, see Figure~\ref{fig:match_pipeline}(c\&d).

\noindent
{\em {4) Refine the Scan Alignment.}} \
Optionally, we can apply iterative closest point (ICP) to refine the alignment generated by the selected optimal transform, see Figure~\ref{fig:match_pipeline}(e).
By comparing Figure~\ref{fig:match_pipeline}(d\&e), we can see that the transform calculated by aligning LRFs is very close to the one refined using ICP.


\begin{figure*}[!b]
  \centering
  \includegraphics[width=0.82\textwidth]{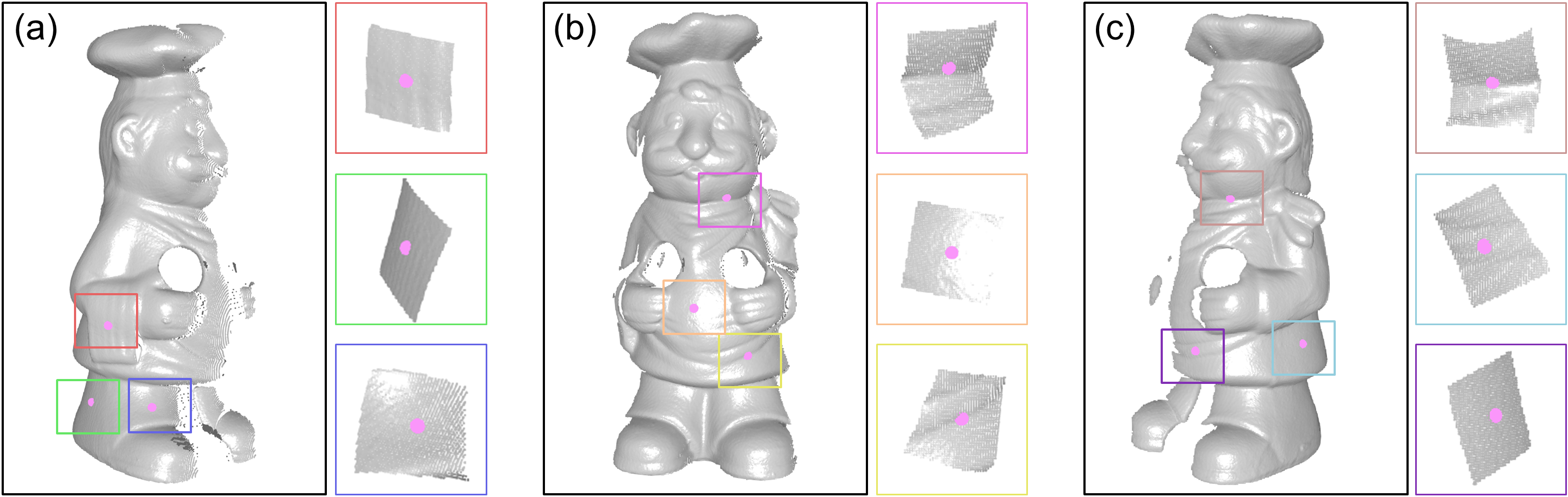}\\
  \caption{Supports on three different scans of a {\sc Chef} model, where feature points are rendered in pink.
   The correspondence between a scan support on the left and its zooming view on the right is indicated by the same 2D box color.}
  \label{fig:desc_saliency}
\end{figure*}

\subsection{Improve Shape Matching using Descriptor Saliency}


To ensure corresponding points to be found on different scans, we need to sample a large number of feature points on each scan, e.g., $M=1000$ in our experiments.
However, among the $M$ descriptors on a single scan, there could exist some descriptors close to one another since their corresponding supports are similar, see Figure~\ref{fig:desc_saliency}(a).
Moreover, among descriptors from all input scans, there could exist a larger number of descriptors with high similarities, see Figure~\ref{fig:desc_saliency}(a-c).


Our observation is that when there exist a large number of descriptors with high similarities, it means their corresponding supports are less distinctive (e.g., flat or spherical shape), see the zooming views in Figure~\ref{fig:desc_saliency}(a).
Thus, it has a lower chance to match the scans correctly by using such supports and their descriptors.
On the other hand, when a descriptor is quite different from others, it means its support is distinctive (see the top zooming views in Figure~\ref{fig:desc_saliency}(b\&c)).


Inspired by this observation, we propose a measure of {\em descriptor saliency} to improve the shape matching performance and compute it based on a descriptor-graph.
The key idea is to find descriptors (and the corresponding supports) that are distinctive by measuring their saliency and apply these descriptors to find corresponding feature points.
We first describe our approach to build a descriptor-graph, present our definition on the descriptor saliency, and then show how we apply the descriptor saliency to enhance shape matching.


\noindent
{\bf Build a Descriptor-Graph.} \
For a given reference scan $\mathrm{S_r}$, we build a descriptor-graph for all the descriptors sampled from $\mathrm{S_r}$ based on their similarities computed using Eq.~\ref{eq:desc_compare}.
Formally, let $\mathrm{G=(V,E)}$ be a descriptor-graph, each node $u \in \mathrm{V}$ represents an SGC descriptor on $\mathrm{S_r}$.
while each directed edge $(u,v)\in \mathrm{E}$ represents that $v$ is one of k-nearest neighbours (k-NN) of $u$ in the descriptor similarity space.
Note that we do not require $u$ also to be one of k-NN of $v$, which means there may not exist a directed edge $(v,u)$ in $\mathrm{G}$.

To build such a graph, a straightforward way is to exhaustive search all descriptors on $\mathrm{S_r}$ to retrieve k-NN for each descriptor in $\mathrm{G}$.
However, this approach is time-consuming, especially when $\mathrm{G}$ is large.
We speed up the creation of the graph following~\cite{Barnes-2010-PatchMatch},
and the basic idea is to initially fill the nearest neighbors by randomly sampling descriptors in $\mathrm{G}$, and iteratively optimize the nearest neighbors locally via similarity propagation and random search until convergence.


\begin{figure*}[!b]
  \centering
  \includegraphics[width=0.45\textwidth]{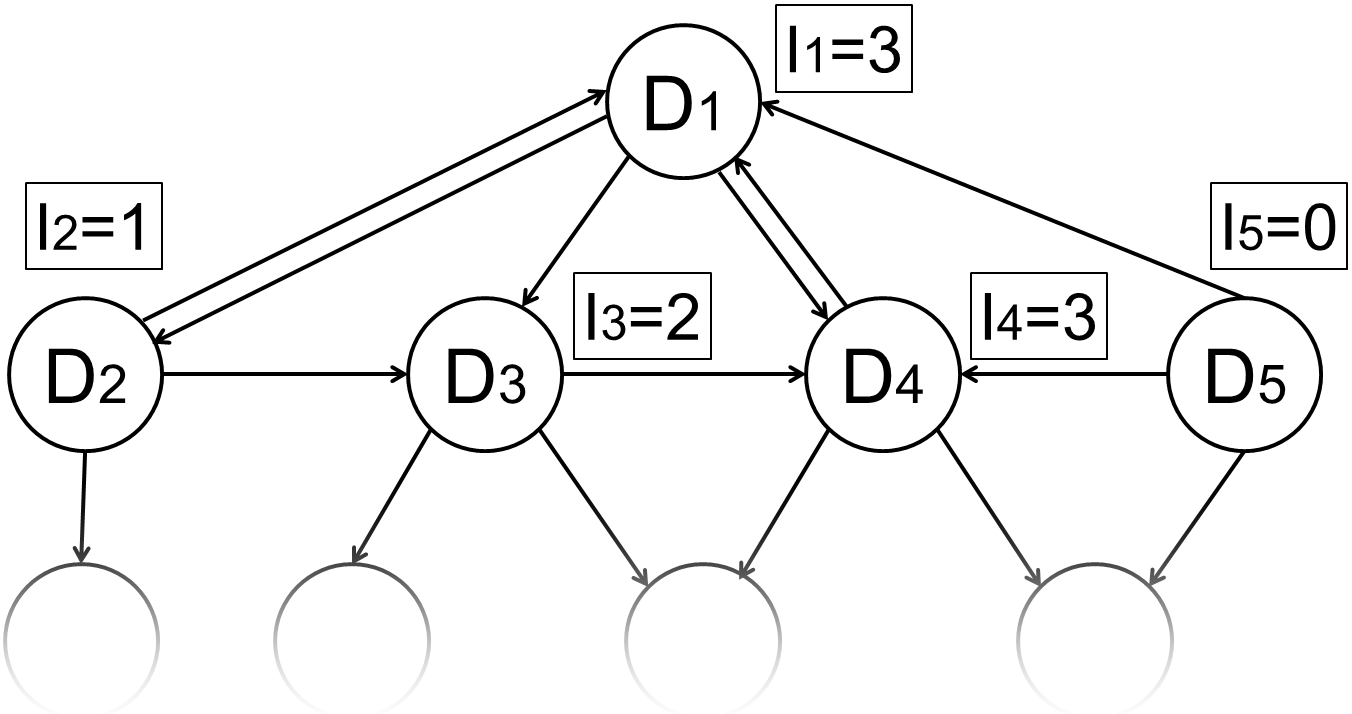}\\
  \caption{An example descriptor-graph (outdegree = 3 for every node).}
  \label{fig:desc_graph}
\end{figure*}

\noindent
{\bf Define Descriptor Saliency.} \
We define descriptor saliency as the distinctiveness among a set of given descriptors.
The larger difference between a descriptor and others, the higher its saliency.
Thus, we measure saliency of a descriptor $\mathrm{D_i}$ in a descriptor-graph $\mathrm{G}$ using $sali(\mathrm{D_i}) = \frac {1}{1+\mathrm{e}^{(I_i- \bar I)}}$,
where $I_i$ denotes the number of nodes in $\mathrm{G}$ that considers $\mathrm{D_i}$ as a k-NN and
$\bar I$ is the mean value of all $I_i$ that is larger than zero.
Note that although
$\mathrm{D_i}$ has $k$ nearest neighbors in $\mathrm{G}$, these neighbors could be very different from $\mathrm{D_i}$.
By fixing $k$, the value $I_i$ can reveal how many descriptors are close to $\mathrm{D_i}$ (i.e., $\mathrm{D_i}$'s distinctiveness).
Figure~\ref{fig:desc_graph} shows descriptor saliency in a simple descriptor-graph with $k=3$.


\noindent
{\bf Shape Matching with Descriptor Saliency.} \
For a given reference scan $\mathrm{S_r}$, we first create a descriptor-graph $\mathrm{G_r}$ for it and compute a saliency value for every descriptor $\mathrm{D_r^i}$ in $\mathrm{G_r}$ using $sali(\mathrm{D_i})$.
For a given descriptor on the data scan $\mathrm{S_d}$, say $\mathrm{D_d^j}$, we enhance the similarity score between $\mathrm{D_d^j}$ and $\mathrm{D_r^i}$ by using $sali(\mathrm{D_r^i})$, i.e., $\bar S(\mathrm{D_d^j}, \mathrm{D_r^i}) = sali(\mathrm{D_r^i})^\alpha \ S(\mathrm{D_d^j}, \mathrm{D_r^i})$,
where $\alpha$ is a weight to control the impact of saliency on the descriptor similarity.
We set $\alpha = 0.2$ in our experiments.

Intuitively, we can find the descriptor on $\mathrm{S_r}$ corresponding to $\mathrm{D_d^j}$ on $\mathrm{S_d}$ by simply comparing every $\mathrm{D_r^i}$ on $\mathrm{S_r}$ with $\mathrm{D_d^j}$ and selecting the one with the largest $\bar S(\mathrm{D_d^j}, \mathrm{D_r^i})$.
We speed up the search of the corresponding descriptor by taking advantage of $\mathrm{G_r}$ with the idea of leveraging existing matches to find better ones.
This is achieved by randomly selecting a set of nodes in $\mathrm{G_r}$ and updating the nodes by a few iterations of similarity propagation and random search~\cite{Gould-2014-SuperpixelGraph}, guided by the similarity score (using Eq.~\ref{eq:desc_compare}) between $\mathrm{D_d^j}$ and the nodes.
After obtaining a small set of descriptors on $\mathrm{S_r}$ that are similar to $\mathrm{D_d^j}$, we conduct re-ranking using $\bar S(\mathrm{D_d^j}, \mathrm{D_r^i})$ to select the final correspondence.

We have illustrated applying descriptor saliency for shape matching between a pair of scans.
Descriptor saliency is more suitable for shape matching among a number of scans, with following changes.
First, we build a large descriptor-graph $\mathrm{G}$ for descriptors from all the scans.
Second, we compare a descriptor on scan $\mathrm{S_m}$ with nodes in $\mathrm{G}$ that are not from $\mathrm{S_m}$.
By this, the larger the number of scans, the higher shape matching performance can be improved by descriptor saliency.


\subsection{Matching Supports Close to Scan Boundary}

Depth scans captured from a certain view are mostly incomplete due to a limited viewing angle, sensor noise, and occlusion.
This results in a surface boundary for a scan.
Matching supports close to the boundary is a challenging task.
First, the support is likely to be incomplete, see examples in Figure~\ref{fig:match_pipeline}(b).
This affects an LRF's repeatability since support is the only input to construct the LRF.
Further, deviation of the LRF affects the construction of the descriptor since support partitioning is performed within the LRF.
Second, the incomplete support directly affects the construction of the descriptor since voxels locating at the missing part(s) become empty, where no shape feature can be extracted.

Due to the above challenges, many existing descriptors are sensitive to the boundary points according to the evaluation in~\cite{Guo-2016-DescriptorEvaluation}.
Therefore, boundary points are usually ignored when applying existing descriptors to partial shape matching~\cite{Mian-2006-ModelBasedRecog,Guo-2013-RotProjection}, assuming that there is sufficient non-boundary scan surface for the matching.
On the other hand, matching boundary points will improve the chance to correctly align different scans, especially when the scan overlap is small.

\begin{figure}[!b]
	\centering
	\includegraphics[width=0.545\textwidth]{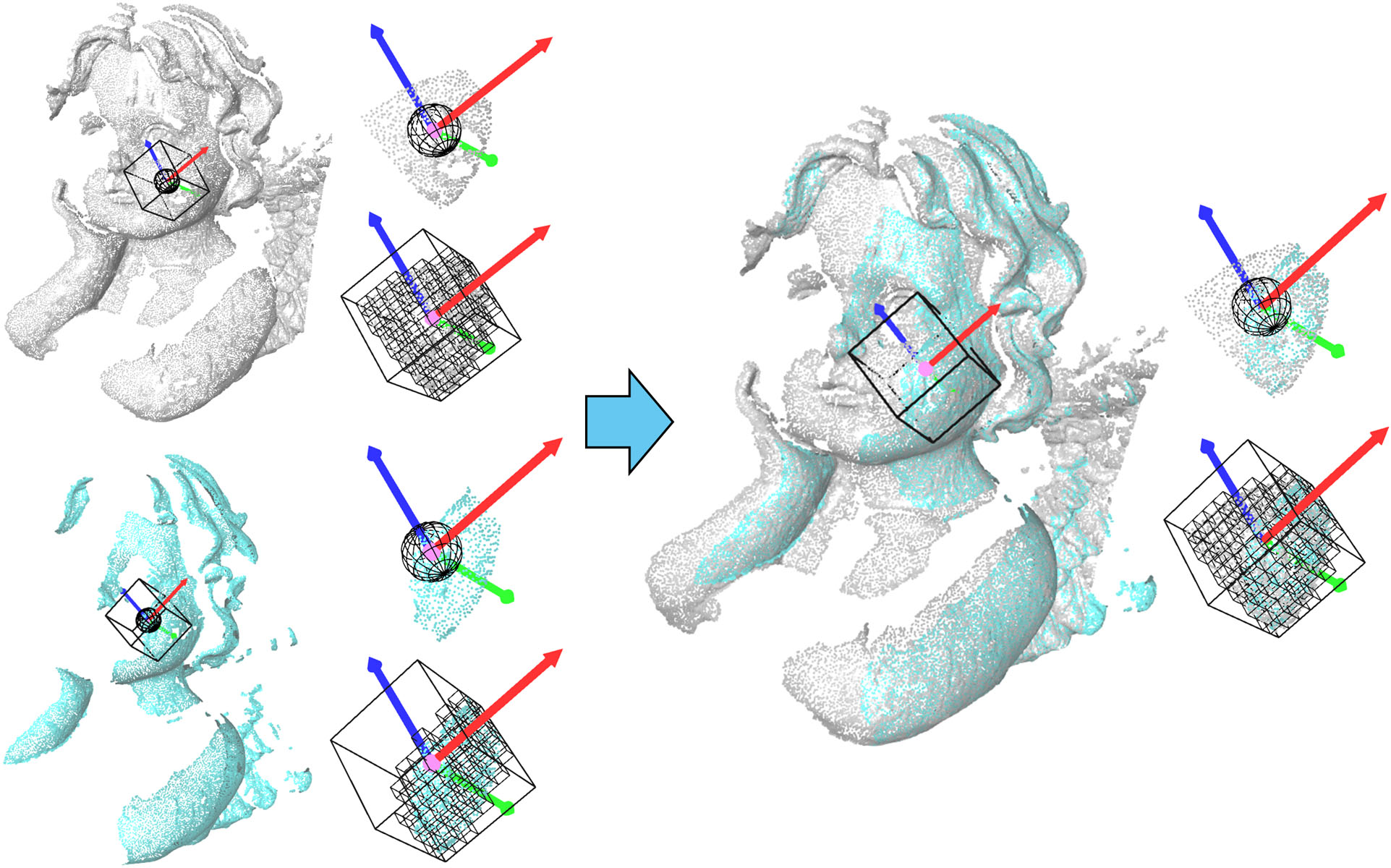}\\
	\caption{(left) Match a support containing holes (in gray scan) with a support close to boundary (in cyan scan) using SGC descriptors;
  (right) aligned scans and supports.}
    \label{fig:match_boundary}
\end{figure}

Our SGC descriptor is especially suitable for handling boundary points for shape matching.
First, the centroid feature that SGC employs is robust against noise and varying point density, which usually happen at scan boundary.
Second, our descriptor comparison scheme allows matching descriptors computed from either a complete or an incomplete support, see Figure~\ref{fig:match_boundary}.
Third, we allow using two different radii for constructing the LRF and the descriptor, i.e., $r{\leq}R$, see supports with varying sizes in Figure~\ref{fig:match_boundary}(left).
By this, a smaller yet complete support can be employed for constructing a repeatable LRF while a larger support allows encoding more (complete or incomplete) local shape for constructing the descriptor.
Based on our experiments, we find that $r=0.5R$ achieves the best performance for matching boundary points when setting $R=20~pr$.

\section{Performance of the SGC Descriptor}
\label{sec:results}

This section evaluates the robustness of SGC with respect to various nuisances, including noise, varying point density, distance to scan boundary, and occlusion.
We compare SGC with three state-of-the-art descriptors that work on point cloud data:
spin image (SI)~\cite{Johnson-1999-SpinImages}, 3DSC~\cite{Frome-2004-3DShapeContext} and SHOT~\cite{Tombari-2010-SHOT}.
Table~\ref{tbl:result_parameter} presents a detailed description of the parameter settings.

\begin{table}[!h]
 \caption{Parameter settings of the four descriptors.}
  \centering
  \includegraphics[width=0.60\textwidth]{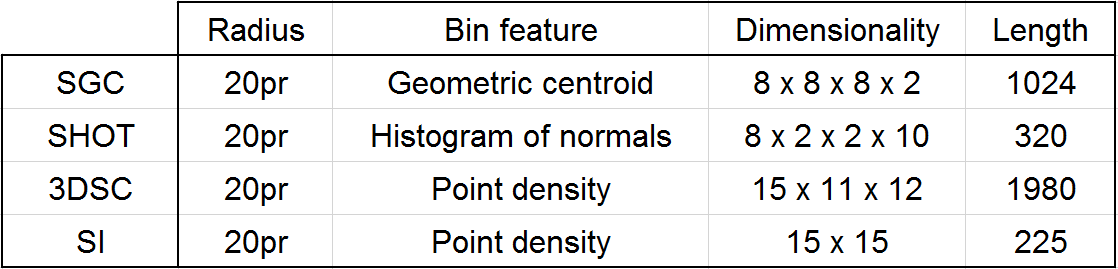}
  \label{tbl:result_parameter}
\end{table}

We perform the experiments on three publicly available datasets: the Bologna dataset~\cite{Tombari-2010-UniqueShapeContext}, UWA dataset~\cite{Mian-2006-ModelBasedRecog}, and Queen's dataset~\cite{Taati-2011-SelectDescriptor}.
Unlike the Bologna dataset that synthesizes complete object models to generate scenes, the scenes in the UWA and Queen's dataset contain partial shape of object models.
We employ the Bologna dataset to evaluate the descriptors' performance with respect to noise and varying point density (Subsection~\ref{subsec:exp_noise}~\&~\ref{subsec:exp_density}),
the UWA dataset to evaluate the descriptors' performance with respect to distance to scan boundary and occlusion (Subsection~\ref{subsec:exp_boundary}~\&~\ref{subsec:exp_occlusion}),
and the Queen's dataset to evaluate improved performance by using descriptor saliency (Subsection~\ref{subsec:saliency}).

We compare the descriptors' performance using RP curves~\cite{Mikolajczyk-2005-RecallPrecision}.
In detail, we randomly select 1000 feature points in each model and find their corresponding points in the scenes via the physical nearest neighbouring search.
By matching the scene features against the model features using each of the four descriptors, an RP curve of the descriptor is generated.


\subsection{Robustness to Noise}
\label{subsec:exp_noise}
\begin{figure*}[!t]
  \centering
  \includegraphics[width=0.93\textwidth]{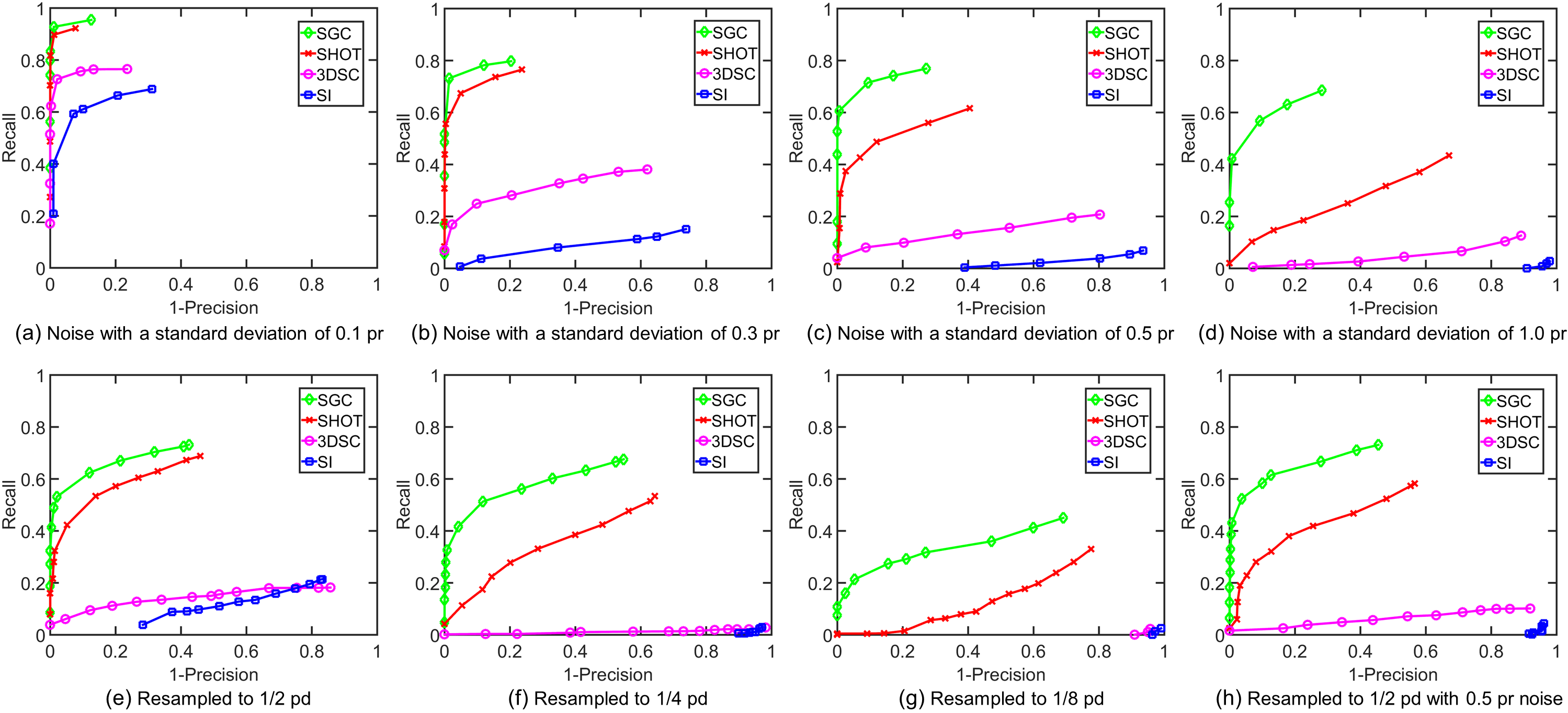}\\
  \caption{RP curves of the four descriptors in the presence of (a-d) noise, (e-g) point cloud downsampling, and (h) their combination.}
  \label{fig:result_noise_density}
\end{figure*}

To evaluate robustness of the descriptors against noise, we add four different levels of Gaussian noise with standard deviations of 0.1, 0.3, 0.5, and 1.0 pr to each scene.
The RP curves of the four descriptors are presented in Figure~\ref{fig:result_noise_density}(a-d).
Thanks to the robust centroid feature, the RP curves show that SGC performs the best under all levels of noise, followed by SHOT and 3DSC.

\subsection{Robustness to Varying Point Density}
\label{subsec:exp_density}
To evaluate robustness of the descriptors with respect to varying point density, we downsample the noise free scenes to 1/2, 1/4 and 1/8 of their original point density (pd).
The RP curves in Figure~\ref{fig:result_noise_density}(e-g) show that SGC outperforms all other descriptors under all levels of downsampling.
Figure~\ref{fig:result_noise_density}(h) shows that SGC performs the best when the input scans are downsampled and contain noise.


\subsection{Robustness to Distance to Scan Boundary}
\label{subsec:exp_boundary}

\begin{figure}[!h]
    \centering
    \includegraphics[width=0.93\textwidth]{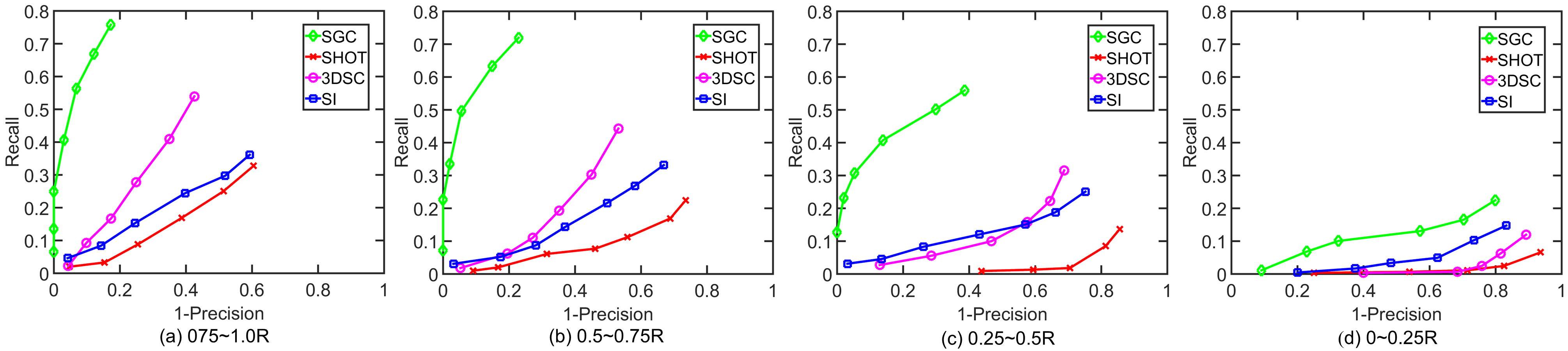}
    \caption{RP curves of feature points in different ranges of distance to the scan boundary.}
    \label{fig:result_boundary}
\end{figure}

We perform experiments for feature points within different ranges of distance to the boundary, i.e., (0, 0.25R], (0.25R, 0.5R], (0.5R, 0.75R], and (0.75R, R].
Note that we set tuned $r=0.5R$ for SGC and $r=R$ for all the other descriptors.
Thanks to the varying support radius and descriptor comparison scheme, Figure~\ref{fig:result_boundary} shows that SGC achieves the best performance for all the four cases.

\subsection{Robustness to Occlusion}
\label{subsec:exp_occlusion}

\begin{figure}[t!]
	\centering
	\includegraphics[width=0.75\textwidth]{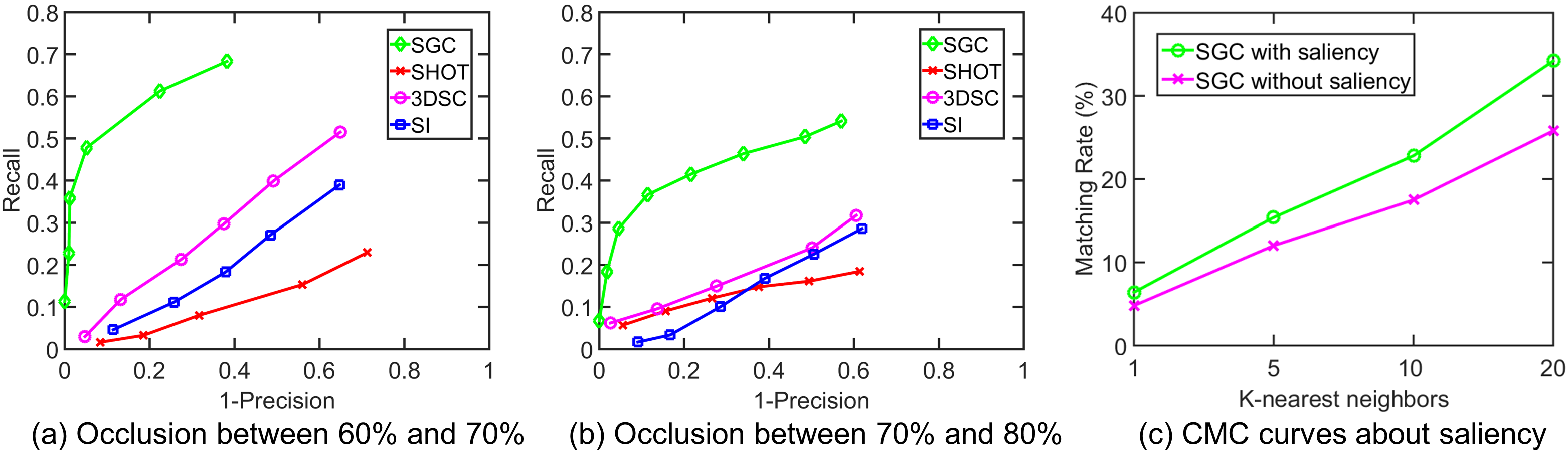}
 \caption{(a\&b) RP curves about occlusion. (c) CMC curves about descriptor saliency.}
  \label{fig:result_occlusion_saliency}
\end{figure}

To evaluate performance of the descriptors under occlusion, we group sampled feature points into two categories following~\cite{Guo-2016-DescriptorEvaluation}, i.e., (60\%, 70\%] and (70\%,80\%] occlusions.
Figure~\ref{fig:result_occlusion_saliency}(a\&b) shows that SGC outperforms all the other descriptors with a large margin since SGC allows handling feature points at scan boundary.

\subsection{Effectiveness of Descriptor Saliency}
\label{subsec:saliency}

To demonstrate effectiveness of descriptor saliency, we compare our shape matching approach with an exhaustive search to find corresponding feature points.
First, we build a descriptor-graph for descriptors sampled from all the five models in the Queen's dataset~\cite{Taati-2011-SelectDescriptor} with $k=16$.
Next, we randomly select 1000 feature points on a scene and calculate their SGC descriptors.
For each scene descriptor, we retrieve its neighbours by searching the descriptor-graph with saliency or exhaustive searching all the model descriptors.
Here, we concern how many neighbours we need to retrieve to ensure the corresponding descriptor is included.
Figure~\ref{fig:result_occlusion_saliency}(c) shows standard Cumulated Matching Characteristics (CMC) curves~\cite{Wang-2007-shape} by using the two approaches.
The curves show that descriptor saliency brings a certain amount of improvement in shape matching.
In addition, descriptor-graph speeds up the search of corresponding descriptors, where each query process takes $0.5 ms$, much faster than the exhaustive search ($62 ms$).


\section{Applications}


\begin{figure*}[!b]
  \centering
  \includegraphics[width=0.60\textwidth]{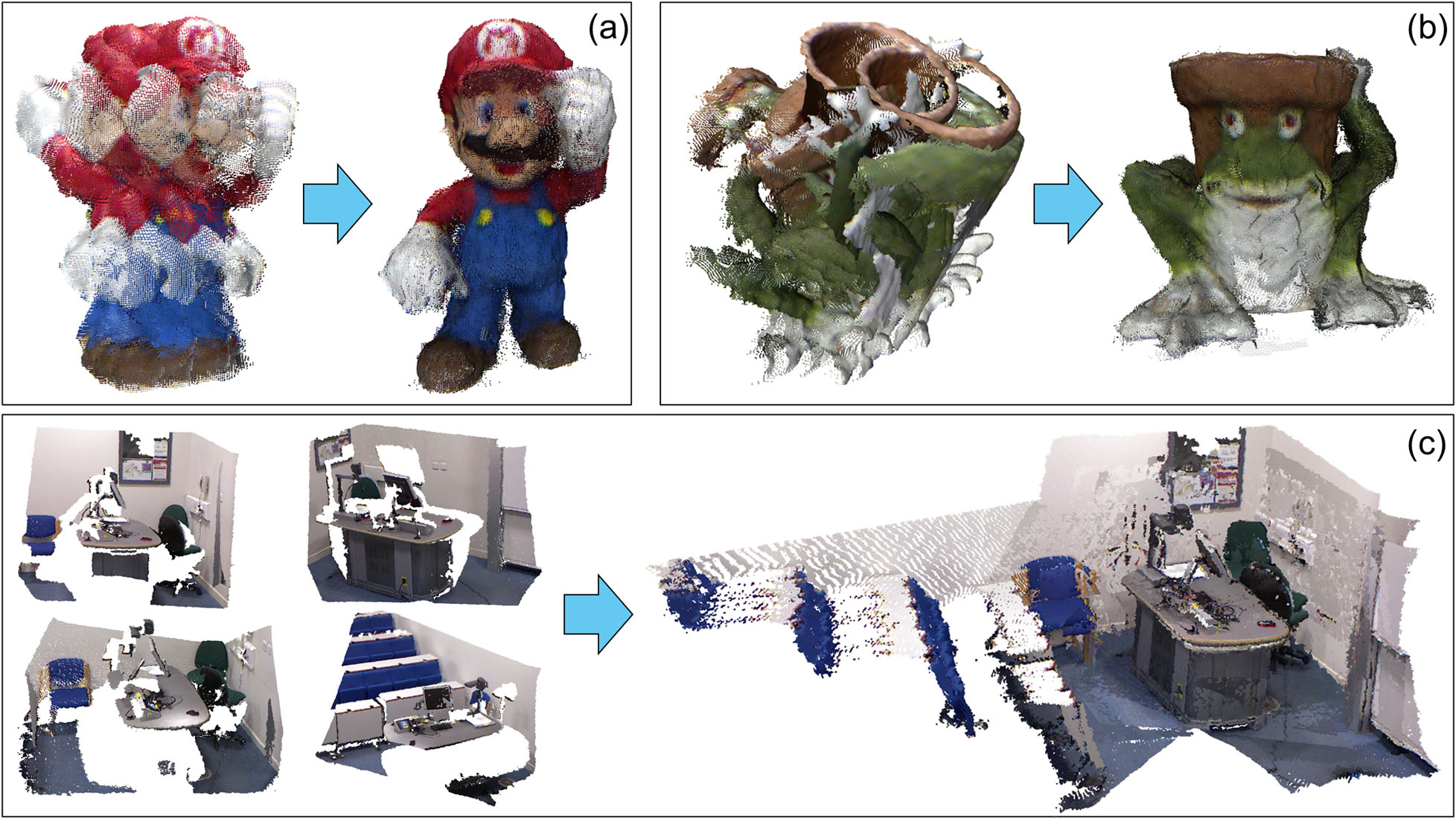}\\
  \caption{ Our reconstruction results. (a) {\sc Super Mario}; (b) {\sc Frog}; and (c) {\sc Stage scene}.}
  \label{fig:app_reconstu_obj_scene}
\end{figure*}

{\bf 3D Object/Scene Reconstruction.} \
To reconstruct a more complete model from a set of scans, we build a descriptor-graph for all the scans.
As the graph has encoded k-NN for each descriptor (and the feature point), we search the corresponding feature point (and its associated scan ID) locally within the k-NN, and align the two scans based on the correspondence and merge them into a larger point cloud.
We keep aligning each of the remaining scans with the point cloud and merging them until all scans are registered.
Figure~\ref{fig:app_reconstu_obj_scene} shows two objects and one scene reconstructed by our approach on different datasets~\cite{Tombari-2010-SHOT,Mellado-2014-Super4PCS}.


\noindent
{\bf 3D Object Recognition.} \
We conduct this experiment on the challenging Queen's dataset~\cite{Taati-2011-SelectDescriptor}.
To represent the model library well with SGC, we remove the noise in each model point cloud and build a descriptor-graph for descriptors sampled from all the models.
For a give scene scan, we also sample a number of SGC descriptors.
By searching a corresponding descriptor in the graph for a given scene descriptor, we know the correspondence between a model in the library and a partial scene, thus recognizing the object in the scene scan.
Note that we recognize a single object at a time and segment the object once recognized.

\begin{figure*}[!h]
  \centering
  \includegraphics[width=0.99\textwidth]{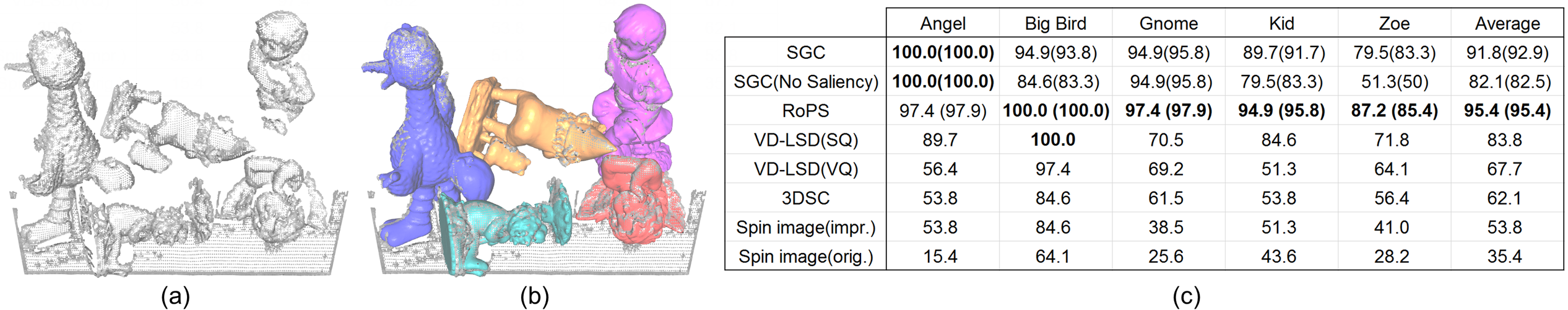}\\
  \caption{Recognition results on the Queen's dataset.
  (a) An example scene;
  (b) our recognition results; and
  (c) recognition rates of the five models (values in brackets are the results on the whole dataset while others are the results on the subset as in~\cite{Taati-2011-SelectDescriptor}).}
  \label{fig:queen_recogn}
\end{figure*}

Figure~\ref{fig:queen_recogn}(a\&b) show the recognition result on an example scene.
Figure~\ref{fig:queen_recogn}(c) shows that SGC based algorithm outperforms most existing methods including VD-LSD~\cite{Taati-2011-SelectDescriptor}, 3DSC~\cite{Frome-2004-3DShapeContext} and spin image~\cite{Johnson-1999-SpinImages} based algorithms.
RoPS based algorithm is the current best 3D object recognition approach and it achieves slighter better performance than SGC with additional mesh information of the scene scans.
In particular, the performance of our algorithm without using descriptor saliency decreases about 10\%, indicating the usefulness of the saliency.

\section{Conclusion}
\label{sec:Conclusion}

We have presented a novel SGC descriptor for matching partial shapes represented by 3D point clouds.
SGC integrates three novel components:
1) a local shape description that encodes robust geometric centroid features;
2) a descriptor comparison scheme that allows comparing supports with missing parts; and
3) a descriptor saliency measure that can identify distinct descriptors.
By this, SGC is robust against various nuisances in point cloud data when performing partial shape matching.
We have demonstrated SGC's performance by comparisons with state-of-the-art descriptors and two partial matching applications.

\section*{Acknowledgments}
This work is supported in part by
the Fundamental Research Funds for the Central Universities (WK0110000044),
Anhui Provincial Natural Science Foundation (1508085QF122),
National Natural Science Foundation of China (61403357, 61175057), and
Microsoft Research Asia Collaborative Research Program.

\bibliographystyle{splncs}
\bibliography{SGC}



\end{document}